\documentclass[conference]{IEEEtran}
\IEEEoverridecommandlockouts

\usepackage[sorting=none]{biblatex} 
\usepackage{amsmath,amssymb,amsfonts}
\usepackage{graphicx}
\usepackage{textcomp}
\usepackage{tabularx}
\usepackage{xcolor}
\usepackage{etoolbox}
\usepackage{multirow}
\usepackage{caption}
\makeatletter
\patchcmd{\@makecaption}
  {\scshape}
  {}
  {}
  {}
\makeatletter
\patchcmd{\@makecaption}
  {\\}
  {.\ }
  {}
  {}
\makeatother
\def\tablename{Table}
\usepackage{algorithm}
\usepackage{algorithmic}
\usepackage{lipsum}
\usepackage{mwe}
\usepackage[section]{placeins}

\def\BibTeX{{\rm B\kern-.05em{\sc i\kern-.025em b}\kern-.08em
    T\kern-.1667em\lower.7ex\hbox{E}\kern-.125emX}}
\begin{document}

\title{Enabling Incremental Training with Forward Pass for Edge Devices
\thanks{\par\noindent\rule{50pt}{0.1pt}\\ *Also at Palo Alto Networks, Santa Clara, CA, USA}}
\author{\IEEEauthorblockN{Dana AbdulQader}
\IEEEauthorblockA{\textit{Electrical Engineering Department} \\
\textit{Santa Clara University}\\
Santa Clara, CA, USA \\}
\and
\IEEEauthorblockN{Shoba Krishnan}
\IEEEauthorblockA{\textit{Electrical Engineering Department} \\
\textit{Santa Clara University}\\
Santa Clara, CA, USA \\
}
\and
\IEEEauthorblockN{Claudionor N. Coelho Jr.*}
\IEEEauthorblockA{\textit{Electrical Engineering Department} \\
\textit{Santa Clara University}\\
Santa Clara, CA, USA \\
}
}


\maketitle

\begin{abstract}
Deep Neural Networks (DNNs) are commonly deployed on end devices that exist in constantly changing environments. In order for the system to maintain it's accuracy, it is critical that it is able to adapt to changes and recover by retraining parts of the network. However, end devices have limited resources making it challenging to train on the same device. Moreover, training deep neural networks is both memory and compute intensive due to the backpropagation algorithm. In this paper we introduce a method using evolutionary strategy (ES) that can partially retrain the network enabling it to adapt to changes and recover after an error has occurred. This technique enables training on an inference-only hardware without the need to use backpropagation and with minimal resource overhead. We demonstrate the ability of our technique to retrain a quantized MNIST neural network after injecting noise to the input. Furthermore, we present the micro-architecture required to enable training on HLS4ML (an inference hardware architecture) and implement it in Verilog. We synthesize our implementation for a Xilinx Kintex Ultrascale Field Programmable Gate Array (FPGA) resulting in less than 1\% resource utilization required to implement the incremental training. 


\end{abstract}


\section{Introduction}



\begin{figure}
    \centering
    \includegraphics[width=8cm]{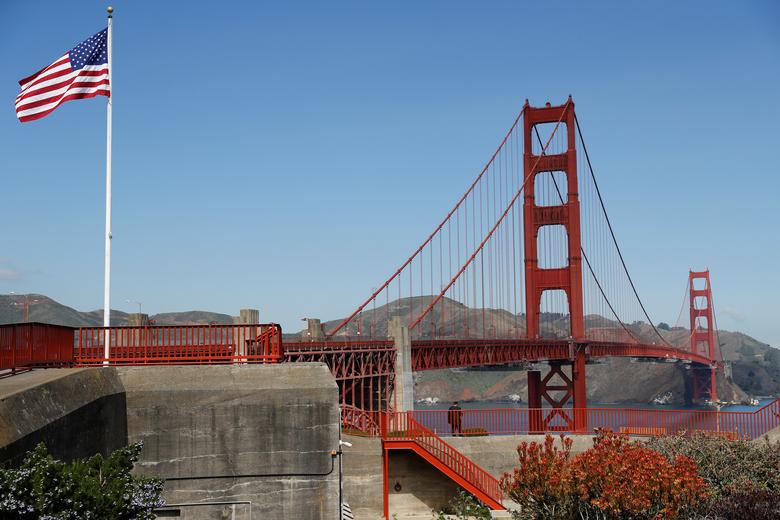}
    \caption{San Francisco before wildfires,  adapted from \cite{SF_before}}
    \label{fig:SF}
\end{figure}

\begin{figure}
    \centering
    \includegraphics[width=8cm]{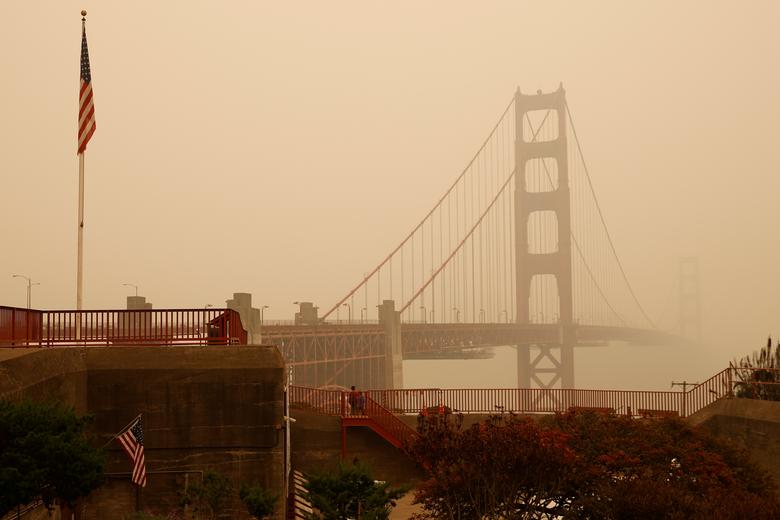}
    \caption{San Francisco during 2020 wildfires, adapted from \cite{SF_after}}
        \label{fig:SF_fire}
\end{figure}

DNN models are usually trained using data that's randomly sampled from a stationary data distribution. Yet, DNN models are often deployed on end devices that exist in environments where the data distribution can shift. This can occur for many reasons such as errors in the sensor capturing the data or environmental changes such as temperature, humidity or pressure. For instance, the model could be trained using images similar to those shown in Figure \ref{fig:SF}. However, during San Francisco 2020 wildfires, the environment changed and so the captured image's statistical properties shifted unexpectedly as can be seen in Figure \ref{fig:SF_fire}. Since machine learning models learn to respond to the probability distribution of the input, as the statistical properties of the input shift the model can no longer make accurate predictions. Hence, if images like those shown in Figure \ref{fig:SF_fire} are fed to the model, it will result in errors. In order to correct this, the model would need to be retrained to this new environment with the new images. Therefore, the ability for a DNN model to adapt to these changes and recover from error is important. This is especially important in situations where the end device is not accessible or in safety-critical applications such as space exploration and autonomous driving \cite{auto2}. 



Retraining a DNN can be challenging on a device with limited hardware resources as training neural networks using backpropagation is a task that is both memory and compute intensive \cite{intensive,deepcompression}. Such training is typically offloaded to the cloud and is almost exclusively done on one or more Graphics Processing Units (GPUs) \cite{GPU}. Nonetheless, being able to retrain on the end device provides many advantages including better privacy, faster processing and less overhead in transferring data \cite{edge}. Moreover, it is especially critical in remote areas where internet connectivity is not possible.  

In this paper, we propose a technique that enables deployed DNN models to adapt and recover from errors by performing incremental training on the device when such errors or shift in the input data occurs. Our method uses an evolutionary strategy (ES) based technique to perform the incremental training without the need to use backpropagation or calculating gradients allowing us to use inference only hardware to perform the incremental training. 

The rest of the paper is divided as follows, section II discusses some of the related work and background information regarding evolution strategy (ES), section III details our methodology, section IV highlights our experiments and results and finally we conclude in section V. 

\section{Background}
Training neural networks, even if partially, is compute and memory intensive as it involves backpropagation and calculating gradients for the backward pass. Many techniques exist that allow inference to be performed in low-precision arithmetic such as in \cite{QNN} and \cite{finn}, yet, high-precision floating-point data types are still required to perform backward propagation. Some research towards lower precision training exists, for instance most recently Sun et al. proposed a new technique that enables 4-bit training \cite{IBM}. Nonetheless, implementing the training using backpropagation even in low-precision would still require the hardware to contain both forward and backward pass micro-architectures on an already resource constrained device. Furthermore, to the best of our knowledge, there has been no implementation of a hardware capable of both inference and training using only the forward pass.  

Few, if any, alternatives exist to backpropagation gradient descent based methods. Salimans et al. recent research results show that a specific type of Evolutionary Strategy (ES) technique can be used for training instead of the gradient-based backpropagation method \cite{OpenAI}. Additionally, research by Zhang et al. further proves the possibility of using Salimans et al. evolutionary strategy technique and demonstrates it's ability to achieve 99\% accuracy on MNIST \cite{Uber} . An evolutionary strategy based technique provides the opportunity to be able to use the forward-pass inference hardware for training instead of needing to add a backward pass architecture. 

Evolutionary Strategy (ES) is a black-box optimization technique inspired by natural evolution. According to Salimans et al., evolutionary strategy optimization works by initially setting the model weights, denoted by \(\omega_t\) at iteration \(t\) where \(0 \leq t \leq k-1\) and \(k\) is the number of iterations, to random variables. A population of \(N\) slightly different weight vectors \(\theta_{t,i}\) are then generated by jittering \(\omega_t\) with Gaussian noise \(\epsilon_{t,i}\) where \(0 \leq i \leq N-1\). The noise vector \(\epsilon_{t,i}\) is sampled  from a Gaussian normal distribution with zero mean and standard deviation \(\sigma\). The loss function \({loss}_{t,i}(\theta_{t,i})\) is then evaluated for each population where:
\begin{equation}
    \theta_{t,i} = \omega_t + \sigma\epsilon_{t,i}
\end{equation}
At each iteration, the gradient can be estimated as

\begin{equation}
    \tilde{g_t} = \frac{1}{N\sigma}\sum_{i=1}^N \epsilon_{t,i}  {loss}_{t,i}
\end{equation}

\noindent the weight parameters are then updated as follows

\begin{equation}
    \omega_{t+1} = \omega_t + \alpha \tilde{g_t}
\end{equation}

\noindent where \(\alpha\) is the learning rate. Mathematically, this is equivalent to estimating the gradient of the loss function in the parameter space using finite differences along N random directions. Hence, evolutionary strategy does not require functions to be differentiable as is the case with gradient descent making it an appealing choice for quantized networks. 

\section{Proposed Methodology}
In this section, we detail our incremental training technique using an evolutionary strategy method. Additionally, we present the hardware micro-architecture required to enable incremental training on an inference-only hardware. Throughout this paper, incremental training is used to refer to our partial training technique where only a select layer is retrained on a pre-trained deployed model. 

\subsection{Algorithm: Incremental training using ES}
Quantized neural networks drastically reduce memory size and can replace most arithmetic operations with bit-wise operations making them popular for hardware deployment. Research has continuously proven to show that despite the low precision, high accuracy is still maintained \cite{QNN}. For those reasons, in our work we focus on using quantized neural networks to ensure minimal required resources to even further enable the ability to add on training. 

\begin{algorithm}
\caption{Incremental Training with Evolutionary Strategy}
\label{Alg:ES1}
\hspace*{\algorithmicindent} \textbf{Input:} {loss function $loss(.)$, vector of weights $\omega$ of size $W$, vector of training data input $x$ of size $M$, vector of training data output $y$ of size $M$, learning rate $\alpha$, noise standard deviation $\sigma$}
\begin{algorithmic}[1]
\FOR{$t = 0,1,2,\ldots , k-1$} 
\FOR{$i = 0,1,2,\ldots , N-1$}
\STATE Sample $\epsilon_{i} \sim \mathcal{N}(0,1)$ where $\epsilon_{i}$ is a vector of size $W$
\STATE Compute $\theta_i = \omega_{i} + \sigma\epsilon_i$ 
\STATE Evaluate $loss_i = loss(forward(x|model, \theta_i), y)$
\ENDFOR
\STATE Estimate gradient $\tilde{g_t}= \frac{1}{N\sigma}\sum_{i=0}^{N-1}\epsilon_i loss_i$
\STATE Set $\omega_{t+1} \leftarrow \omega_t + \alpha \tilde{g_t} $
\ENDFOR
\end{algorithmic}
\end{algorithm}

Given a trained quantized neural network, a new set of training data that represents the shift in the data distribution is used to incrementally train the model and improve its accuracy. It is important to note that since we are performing incremental training, the size of the training data can be much smaller than the size of the training data used to initially train the network. Algorithm \ref{Alg:ES1} demonstrates our method, we use Salimans et al. evolutionary strategy technique described in the previous section to perform the training. In our algorithm, a subset of the model weights or the weights of a specific layer are chosen to be trained. At each iteration the loss function is evaluated for each generated population over the entire training data. The loss function used in our method is the negative of the mean absolute error (MAE). This was chosen as it is computationally less expensive than other loss functions such as mean square error (MSE) or root mean square error (RMSE). No multiplication or square root is involved making MAE more feasible to implement on a minimal resource device.

\subsection{Incremental Training Micro-Architecture }
\begin{figure}
   \centering
    \includegraphics[width=6cm]{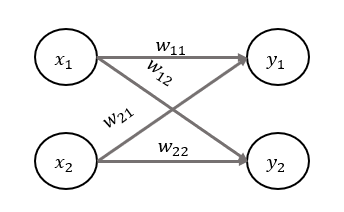} 
    \caption{Simple Neural Network}
    \label{fig:NN}
\end{figure}

\begin{figure}
    \centering
    \includegraphics[width=8cm]{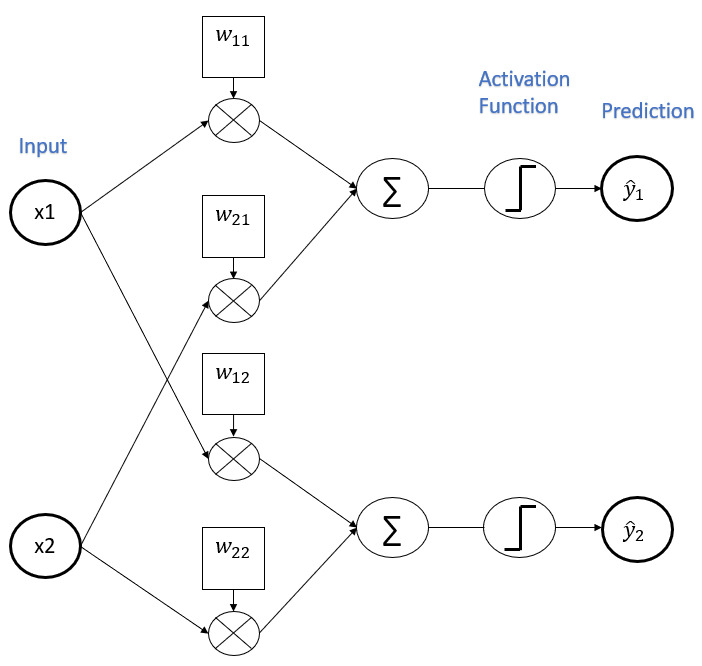} 
    \caption{Micro-architecture of a simple neural network using HLS4ML}
    \label{fig:Arch_NN}
\end{figure}

Implementing an evolutionary strategy based technique allows us to utilize the existing forward pass micro-architecture for training by making small adjustments to it. For instance, HLS4ML is a compiler package that translates a neural network model into an HLS project that can be implemented to run inference on an FPGA \cite{HLS4ML}. In HLS4ML, the inference is pipelined where each layer of the neural network is one stage of the pipeline allowing inputs to be streamed after it's initiation interval. Consider the simple network illustrated in Figure \ref{fig:NN} with 1 input layer and 1 output layer where each layer has 2 neurons. Figure \ref{fig:Arch_NN} demonstrates the micro-architecture of the simple fully connected neural network model if implemented using HLS4ML. On such a model, adding a backpropagation based technique for training would require a different micro-architecture \cite{book}. 


\begin{figure}
    \centering
    \includegraphics[width=8cm]{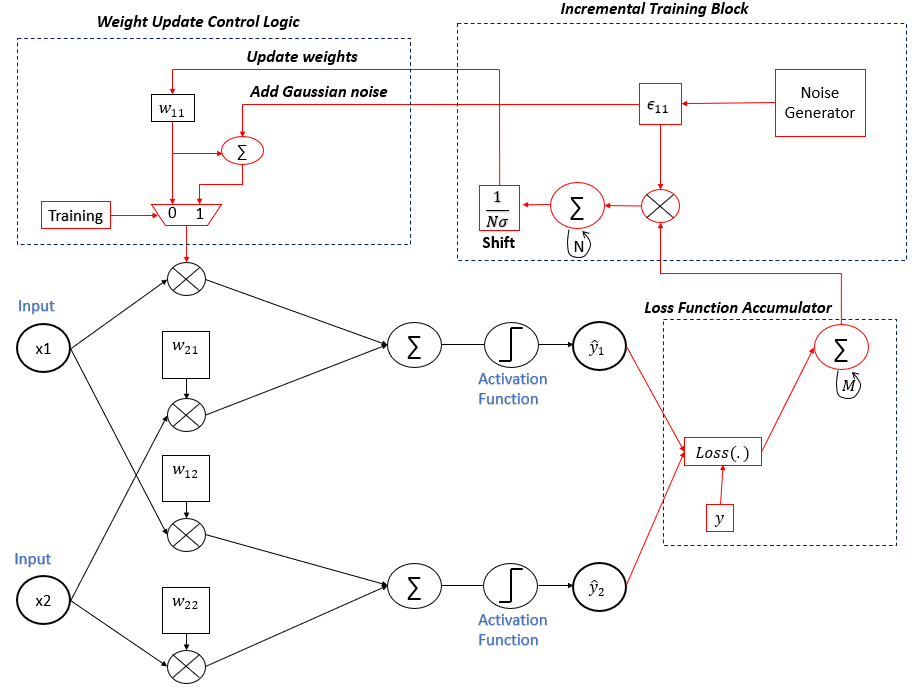}
    \caption{Incremental training micro-architecture with forward pass}
   \label{fig:Arch_ES}
\end{figure}

Figure \ref{fig:Arch_ES} shows the minor adjustments to the  micro-architecture that would need to be made in order to allow for incremental training using our technique where \(M\) denotes the number of training images. The Incremental Training block along with the Weight Update Control Logic can be reused to update other weights or can be duplicated for each weight intended to be updated during training. Reusing the same block allows for further reduction in the required resources but would require more time to complete the training. Also, it is important to note that the weights that need to be updated for training can not be fused in the logic and instead would need to exist in BRAMs. Since only a fraction of the total weights will be updated for training, the number of BRAMs required is feasible.

The Weight Update Control Logic block works such that it allows for a round of inference, followed by a round of training and then back to inference if needed. Training data enters as input, and a training signal triggers the weight update control logic to enable training. If inference is needed, the training operation can be paused without losing any of the training effort and inference can be resumed.The device does not need to halt it's operation as it waits for training to be complete, instead training iterations can occur in between different inference passes until training is complete.  Of course, this will mean training will take more time but it allows for close to continuous operation of the device.


\section{Experiments and Results}
In this section we outline our results using our method to retrain an MNIST Neural Network and show it's ability to improve on it's accuracy after being injected with noise. Furthermore, we implement our solution in Verilog and synthesize it for a Xilinx Ultrascale FPGA and demonstrate the minimal resources required for our technique. 

\subsection{Incremental Training on MNIST}
\begin{figure}[!htb]
    \centering
    \includegraphics[width=8cm]{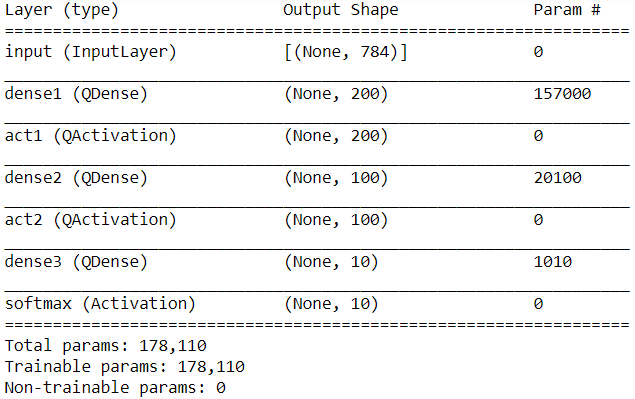}
    \caption{QKeras MNIST Network}
    \label{fig:MNIST}
\end{figure}

We implemented our training method using QKeras and Tensorflow and experimented with MNIST dataset. MNIST is an image classification benchmark dataset \cite{MNIST}. It consists of a training set of 60K and a test set of 10K 28x28 gray-scale images representing digits ranging from 0 to 9. The QKeras neural network we trained on MNIST consists of 3 hidden layers with 178,110 total trainable parameters. QKeras is an extension of the Keras library that allows for the creation of quantized deep neural network models. Qkeras models are trained quantization-aware allowing for lower precision while still maintaining accuracy and in return significantly reducing resource consumption when implemented on an FPGA \cite{QKeras}. The neural network was implemented in 4 bits fixed point arithmetic and it's summary is show in Figure \ref{fig:MNIST}.

\begin{figure}[!htb]
    \centering
    \includegraphics[width=8cm]{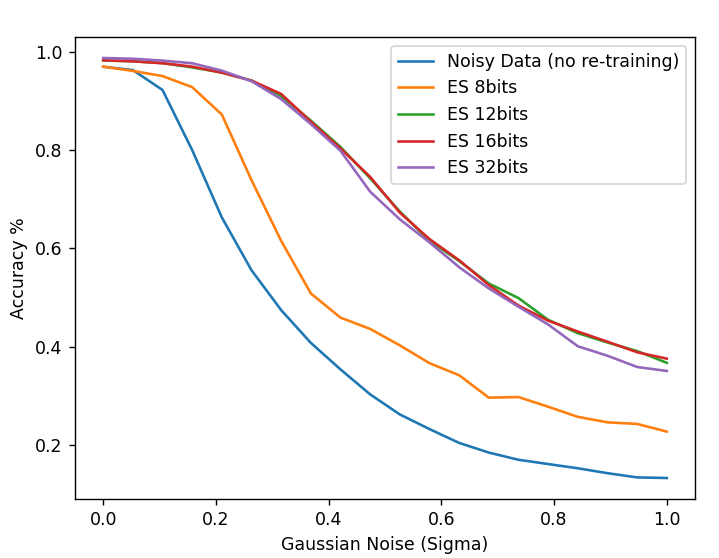}
    \caption{Graph showing improved accuracy after retraining using our incremental training technique at different quantization levels.}
    \label{fig:graph_trained}
\end{figure}

We will now describe the steps we performed for our experimentation. Step 1: we split the MNIST training dataset to 50,000 for training and 10,000 for validation. Step 2:  we trained the Qkeras network using the training dataset - this represents the deployed mode. Step 3:  we injected the validation dataset with Gaussian noise levels ranging from \(0 \leq \sigma \leq 1\) to create a noisy dataset. This is to model a fault in the input or sensor of the system. Since neural network models respond to the probability distribution of the input, a change in the input or an error in the sensor capturing the input cases a shift in it's probability distribution as we illustrated in Figures \ref{fig:SF} and \ref{fig:SF_fire}. Hence, injecting noise into the input would model a similar behavior. Step 4: we evaluated the model with the noisy dataset - results are shown in blue in Figure \ref{fig:graph_trained}. As expected, when the noise level increases the models ability to make accurate prediction decreases. Step 5: we retrained the first hidden layer of the model using our technique and the noisy dataset. Step 5 demonstrates the models ability to learn and improve it's accuracy after retraining. Lastly, step 6: we repeated steps 4 and 5 using different data precision for the incremental training portion. 

For all our experiments, we used 100 for the population size N and 100 for the number of iterations k.  Zhang et al. results show that to fully train an MNIST network, a population size of  10,000 and 2,000 iterations were required \cite{Uber}. However, since we are performing incremental training on an already trained network, a much smaller number for both the population size and number of iterations can be used. Results are shown in the graph in Figure \ref{fig:graph_trained}. The `Noisy Data' curve shows how the accuracy of a model can drop as the statistical properties of the input change, i.e. the deployed model evaluated with the noisy dataset as input. The `ES 32bits' curve represents the model's accuracy after it's been retrained using our incremental training ES based technique and the noisy dataset. As can be seen, using our ES based method we were able to train the network and improve it's accuracy.

The graph in Figure \ref{fig:graph_trained} also shows our results when using different fixed point representations  for our incremental training. It can be observed that the precision can be reduced from 32 bits to 12 bits without causing a loss in accuracy. The ability to train in lower precision allows us to further reduce the required resources as compared to floating point arithmetic while maintaining accuracy. 




ES techniques perform gradient estimation using random sampling. Because of that, the sample size needs to be large enough for the estimation to be valid. DNN inference time on specialized hardware can be extremely fast and efficient, Coelho et al. presented a case study using QKeras and HLS4ML where they were able to achieve an inference speed of 1$\mu$\(s\)  \cite{QKeras}. This specific case study involves classifying hundreds of terabytes of data from proton-proton collisions at the CERN Large Hadron Collider using a Qkeras DNN with 3 hidden layers (64, 32 and 32 nodes, respectively). Since fast inference with a latency of $\mathcal{O}(1\mu s)$ can be achieved on an FPGA, we can expect to be able to achieve very fast training even with large number of iterations. The time required for a single training iteration using one training image to be complete can be estimated as follows:
\begin{equation}
    Training\_Time = (t_{f} + t_{l} + t_{g} +t_{u})\frac{W}{P}N 
\end{equation}

\noindent where \(W\) is the number of weights to be updated, \(P\) is the number of incremental training blocks and weight update control logic implemented, \(t_f\) is inference time, \(t_l\) is the time to calculate the loss, \(N\) is the population size, and finally \(t_g\) and \(t_u\) is the time required to calculate the gradient and update the weights respectively. 

In our experiments, we performed exponent quantization of the loss. This allows for efficient multiplication as the numbers are quantized to a power-of-2 representation, and the multiplication can be achieved by shift \cite{QKeras}. As a result, the multiplier in Figure \ref{fig:Arch_ES} can be replaced with a programmable shift register. Given the very minimal hardware involved, \(t_{l}\), \(t_{g}\) and \(t_{u}\) are negligible. Inference time and the number of incremental training blocks implemented are the main factors in determining total training time. Hence, the overall training time can be estimated as: 

\begin{equation}
\label{eq:t}
    Training\_Time =  t_f\frac{W}{P}MNk 
\end{equation}

\noindent where \(M\) is the number of images, N is the population size and k is the number of iterations.If both the population size \(N\) and the number of iterations \(k\) equals to 100 and P is equal to the number of weights, using equation \ref{eq:t} and assuming \(1\mu s\) for \(t_f\), the training can be done in approximately 100 seconds for a training dataset  size \(M\) of 10,000 images. The incremental training can even be mixed with inference jobs without having to halt inference for the entire training time. 

\subsection{FPGA Resource Utilization}

We implemented our incremental training design in Verilog and synthesized it for a Xilinx Kintex Ultrascale with part number xcku035-fbva676-3-e. For our design we used an 8-bit Linear Feedback Shift Register (LFSR) for the noise generator and 32-bit accumulators for estimating the gradient and accumulating the loss. The implementation of the weight update logic involves storing the subset of weights to be trained in BRAMs, this part of the design was not included in our Verilog implementation. 

\begin{table}[ht]
\begin{tabular}{|l|l|l|l|l|}
\hline
Block                                                                                  & Resource & Utilization & Available & Utilization \% \\ \hline
\multirow{2}{*}{\begin{tabular}[c]{@{}l@{}}Incremental \\ Training Block\end{tabular}} & LUT      & 91          & 203128    & 0.04           \\ \cline{2-5} 
                                                                                       & FF       & 68          & 406256    & 0.017          \\ \hline
\multirow{2}{*}{\begin{tabular}[c]{@{}l@{}}Loss Function \\ Accumulator\end{tabular}}  & LUT      & 9           & 203128    & 0.01           \\ \cline{2-5} 
                                                                                       & FF       & 37          & 406256    & 0.01           \\ \hline
\end{tabular}
\caption{Resource utilization estimates for the incremental training micro-architecture}
\label{resource}
\end{table}

The FPGA resource utilization required for the incremental training micro-architecture is shown in Table \ref{resource}. The table represents the number of FFs and LUTs required to implement our Incremental Training Block and the Loss Function Accumulator on any forward-pass architecture.

In our MNIST experiments, we retrained our first layer which has 157,000 trainable parameters as shown in Figure \ref{fig:MNIST}. Table \ref{table:2} shows the area overhead required to update all 157,000 parameters for different values of P, where P is the number of implemented Incremental Training Blocks. Additionally, it presents the training time required in number of forward passes per iteration per image to train all 157,000 parameters for a population size N of 100. As mentioned earlier, Coelho et al. shows that it is possible to achieve inference of $1\mu s$ on an FPGA. If 1000 Incremental Training Blocks are implemented, we can see that it can be possible to achieve a training time per iteration per image of 15.7ms. 


\begin{table}[ht]
\centering
    \begin{tabular}{|c|l|c|c|}
    \hline
    \multirow{2}{*}{\bfseries P} & 
    \multirow{2}{*}{\bfseries Training Time*} & 
    \multicolumn{2}{|c|}{\bfseries Area Overhead}\\ \cline{3-4}
    && LUT & FF \\ \hline
    1 & 15,700,000 & 91 & 68\\ \hline
    10 & 1,570,000 & 910 & 680\\ \hline
    100 & 157,000 & 9,100 & 6,800 \\ \hline
    1k & 15,700 &91,000 & 68,000 \\ \hline
    2k & 7,850 & 182,000 & 136,000\\ \hline
    \end{tabular}
 \caption{The table present the incremental training time and area overhead with varying P. * The training time is in terms of number of forward passes per iteration per image.}
 \label{table:2}
\end{table}


\section{Conclusion}
In this work, we propose an incremental training technique using ES that enables training DNNs on an inference only hardware without the need to use backpropagation. Our work can be expanded further to enable training Convolutional Neural Networks (CNNs) as well. This allows the neural network model to adapt to changes and recover from errors. We implemented our technique using low-precision fixed point arithmetic to further reduce the required resources and demonstrated our techniques effectiveness by showing it's ability to recover after injecting noise into the input. Furthermore, we implemented our method in Verilog and synthesized it for an Ultrascale Xilinx FPGA demonstrating the minimal resources required to implement this training technique. 

\printbibliography 

@article{auto2,
  title={An end-to-end deep neural network for autonomous driving designed for embedded automotive platforms},
  author={Koci{\'c}, Jelena and Jovi{\v{c}}i{\'c}, Nenad and Drndarevi{\'c}, Vujo},
  journal={Sensors},
  volume={19},
  number={9},
  pages={2064},
  year={2019},
  publisher={Multidisciplinary Digital Publishing Institute}
}

@article{edge,
  title={Embedded deep neural network processing: Algorithmic and processor techniques bring deep learning to iot and edge devices},
  author={Verhelst, Marian and Moons, Bert},
  journal={IEEE Solid-State Circuits Magazine},
  volume={9},
  number={4},
  pages={55--65},
  year={2017},
  publisher={IEEE}
}

@article{QNN,
  title={Quantized neural networks: Training neural networks with low precision weights and activations},
  author={Hubara, Itay and Courbariaux, Matthieu and Soudry, Daniel and El-Yaniv, Ran and Bengio, Yoshua},
  journal={The Journal of Machine Learning Research},
  volume={18},
  number={1},
  pages={6869--6898},
  year={2017},
  publisher={JMLR. org}
}

@article{IBM,
  title={Ultra-Low Precision 4-bit Training of Deep Neural Networks},
  author={Sun, Xiao and Wang, Naigang and Chen, Chia-Yu and Ni, Jiamin and Agrawal, Ankur and Cui, Xiaodong and Venkataramani, Swagath and El Maghraoui, Kaoutar and Srinivasan, Vijayalakshmi Viji and Gopalakrishnan, Kailash},
  journal={Advances in Neural Information Processing Systems},
  volume={33},
  year={2020}
}

@article{OpenAI,
  title={Evolution strategies as a scalable alternative to reinforcement learning},
  author={Salimans, Tim and Ho, Jonathan and Chen, Xi and Sidor, Szymon and Sutskever, Ilya},
  journal={arXiv preprint arXiv:1703.03864},
  year={2017}
}

@article{Uber,
  title={On the relationship between the openai evolution strategy and stochastic gradient descent},
  author={Zhang, Xingwen and Clune, Jeff and Stanley, Kenneth O},
  journal={arXiv preprint arXiv:1712.06564},
  year={2017}
}

@article{deepcompression,
  title={Deep compression: Compressing deep neural networks with pruning, trained quantization and huffman coding},
  author={Han, Song and Mao, Huizi and Dally, William J},
  journal={arXiv preprint arXiv:1510.00149},
  year={2015}
}

@article{QKeras,
  title={Ultra Low-latency, Low-area Inference Accelerators using Heterogeneous Deep Quantization with QKeras and hls4ml},
  author={Coelho Jr, Claudionor N and Kuusela, Aki and Zhuang, Hao and Aarrestad, Thea and Loncar, Vladimir and Ngadiuba, Jennifer and Pierini, Maurizio and Summers, Sioni},
  journal={arXiv preprint arXiv:2006.10159},
  year={2020}
}

@article{finn,
  title={FINN-R: An end-to-end deep-learning framework for fast exploration of quantized neural networks},
  author={Blott, Michaela and Preu{\ss}er, Thomas B and Fraser, Nicholas J and Gambardella, Giulio and O’brien, Kenneth and Umuroglu, Yaman and Leeser, Miriam and Vissers, Kees},
  journal={ACM Transactions on Reconfigurable Technology and Systems (TRETS)},
  volume={11},
  number={3},
  pages={1--23},
  year={2018},
  publisher={ACM New York, NY, USA}
}

@article{HLS4ML,
  title={Fast inference of deep neural networks in FPGAs for particle physics},
  author={Duarte, Javier and Han, Song and Harris, Philip and Jindariani, Sergo and Kreinar, Edward and Kreis, Benjamin and Ngadiuba, Jennifer and Pierini, Maurizio and Rivera, Ryan and Tran, Nhan and others},
  journal={Journal of Instrumentation},
  volume={13},
  number={07},
  pages={P07027},
  year={2018},
  publisher={IOP Publishing}
}

@article{intensive,
  title={Low-memory neural network training: A technical report},
  author={Sohoni, Nimit Sharad and Aberger, Christopher Richard and Leszczynski, Megan and Zhang, Jian and R{\'e}, Christopher},
  journal={arXiv preprint arXiv:1904.10631},
  year={2019}
}

@inproceedings{GPU,
  title={Deep learning with COTS HPC systems},
  author={Coates, Adam and Huval, Brody and Wang, Tao and Wu, David and Catanzaro, Bryan and Andrew, Ng},
  booktitle={International conference on machine learning},
  pages={1337--1345},
  year={2013},
  organization={PMLR}
}

@book{SF_before, title={The Golden Gate Bridge}, url={https://www.reuters.com/news/picture/scenes-from-a-smoky-san-francisco-before-idUSRTX7UXXL}, journal={reuters.com}, author={Lam/ REUTERS, Stephen}, year={2020}, month={03}}

@book{SF_after, title={The Golden Gate Bridge is seen under a smoke filled sky from California wildfires in San Francisco, California}, url={https://www.reuters.com/news/picture/scenes-from-a-smoky-san-francisco-before-idUSRTX7UXXL}, journal={reuters.com}, author={Lam/ REUTERS, Stephen}, year={2020}, month={09}}

@article{MNIST,
  author = {LeCun, Yann and Cortes, Corinna},
  howpublished = {http://yann.lecun.com/exdb/mnist/},
  lastchecked = {2016-01-14 14:24:11},
  title = {{MNIST} handwritten digit database},
  url = {http://yann.lecun.com/exdb/mnist/},
  username = {mhwombat},
  year = 1998
}

@book{book,
  title={Deep Learning from Scratch: Building with Python from First Principles},
  author={Weidman, Seth},
  year={2019},
  publisher={" O'Reilly Media, Inc."}
}
\end{document}